\documentclass[11pt,a4paper]{article}
\usepackage[hyperref]{acl2019}
\usepackage{times}
\usepackage{latexsym}
\usepackage{amsmath}
\usepackage{amssymb}
\usepackage{mathtools}
\usepackage{bm}
\usepackage{graphicx}
\usepackage{subcaption}
\usepackage{scrextend}
\usepackage{url}
\usepackage[normalem]{ulem}

\usepackage{algorithm}
\usepackage[noend]{algpseudocode}

\makeatletter
\newcommand{\algmargin}{\the\ALG@thistlm}
\makeatother
\newlength{\whilewidth}	
\algdef{SE}[parWHILE]{parWhile}{EndparWhile}[1]
{\parbox[t]{\dimexpr\linewidth-\algmargin}{%
		\hangindent\whilewidth\strut\algorithmicwhile\ #1\ \algorithmicdo\strut}}{\algorithmicend\ \algorithmicwhile}%
\algnewcommand{\parState}[1]{\State%
	\parbox[t]{\dimexpr\linewidth-\algmargin}{\strut #1\strut}}

\aclfinalcopy %
\def\aclpaperid{359} %

\DeclareMathOperator*{\argmax}{argmax}

\title{A Cross-Sentence Latent Variable Model for \\ Semi-Supervised Text Sequence Matching}

\author{Jihun Choi, \enskip Taeuk Kim, \enskip Sang-goo Lee \\
	Department of Computer Science and Engineering \\
	Seoul National University \\
	{\tt \{jhchoi,taeuk,sglee\}@europa.snu.ac.kr}
}

\date{}
\begin{document}
	\maketitle
	\begin{abstract}
        We present a latent variable model for predicting the relationship between a pair of text sequences.
        Unlike previous auto-encoding--based approaches that consider each sequence separately, our proposed framework utilizes both sequences within a single model by generating a sequence that has a given relationship with a source sequence.
        We further extend the cross-sentence generating framework to facilitate semi-supervised training.
        We also define novel semantic constraints that lead the decoder network to generate semantically plausible and diverse sequences.
        We demonstrate the effectiveness of the proposed model from quantitative and qualitative experiments, while achieving state-of-the-art results on semi-supervised natural language inference and paraphrase identification.
	\end{abstract}
	
	\section{Introduction}
	\label{sec:intro}
	Text sequence matching is a task whose objective is to predict the degree of match between two or more text sequences.
	For example, in natural language inference, a system has to infer the relationship between a premise and a hypothesis sentence,
	and in paraphrase identification a system should find out whether a sentence is a paraphrase of the other.
	Since various natural language processing problems, including answer sentence selection, text retrieval, and machine comprehension, involve text sequence matching components,
	building a high-performance text matching model plays a key role in enhancing quality of systems for these problems \cite{tan2016qa,rajpurkar2016squad,wang2017compare,tymoshenko2018crosspair}.
	
	With the emergence of large-scale corpora, end-to-end deep learning models are achieving remarkable results on text sequence matching;
	these include architectures that are linguistically motivated \cite{bowman2016spinn,chen2017esim,kim2019dynamic},
	that introduce external knowledge \cite{chen2018kim},
	and that use attention mechanisms \cite{parikh2016decomposable,shen2018disan}.
	The recent deep neural network--based work on text matching could roughly be categorized into two subclasses:
	i) methods that exploit inter-sentence features
	and
	ii) methods based on sentence encoders.
	In this work, we focus on the latter where sentences%
	\footnote{%
		Throughout the paper, we will use the term `sequence' and `sentence' interchangeably unless ambiguous.
	}
	are separately encoded using a shared encoder and then fed to a classifier network,
	due to its efficiency and general applicability across tasks.
	
	Meanwhile, despite the success of deep neural networks in natural language processing,
	the fact that they require abundant training data might be problematic,
	as constructing labeled data is a time-consuming and labor-intensive process.
	To mitigate the data scarcity problem, several semi-supervised learning paradigms, that take advantage of unlabeled data when only some of the data examples are labeled \cite{chapelle2010semisupervised}, are proposed.
	These unlabeled data are much easier to collect, thus utilizing them could be a good option;
	for example in text matching, possibly related sentence pairs could be retrieved from a database of text via simple heuristics such as word overlap.
	
	In this paper, we propose a cross-sentence latent variable model for semi-supervised text sequence matching.
	The proposed framework is based on deep probabilistic generative models \cite{kingma2014vae,rezende2014dgm} and is extended to make use of unlabeled data.
    As it is trained to generate a sentence that has a given relationship with a source sentence, both sentences in a pair are utilized together, and thus training objectives are defined more naturally than other models that consider each sentence separately \cite{zhao2018arae,shen2018lvm}.
	To further regularize the model to generate more plausible and diverse sentences, we define semantic constraints and use them for fine-tuning.
	
	From experiments, we empirically prove that the proposed method significantly outperforms previous work on semi-supervised text sequence matching.
	We also conduct extensive qualitative analyses to validate the effectiveness of the proposed model.
	
	The rest of the paper is organized as follows.
	In \S\ref{sec:background}, we briefly introduce the background for our work.
	We describe the proposed cross-sentence latent variable model in \S\ref{sec:framework},
	and give results from experiments in \S\ref{sec:experiments}.
	We study the prior work related to ours in \S\ref{sec:related-work} and conclude in \S\ref{sec:conclusion}.
	
	\section{Background}
	\label{sec:background}
	
	\subsection{Variational Auto-Encoders}
	\label{subsec:vae}
	Variational auto-encoder (VAE, \citealp{kingma2014vae}) is a deep generative model for modeling the data distribution $p_{\bm\theta}(\mathbf{x})$.
	It assumes that a data point $\mathbf{x}$ is generated by the following random process:
	(1) $\mathbf{z}$ is sampled from $p (\mathbf{z})$ and
	(2) $\mathbf{x}$ is generated from $p_{\bm\theta} (\mathbf{x}|\mathbf{z})$.
	
	Thus the natural training objective would be to directly maximize the marginal log-likelihood $\log p_{\bm\theta} (\mathbf{x}) = \log \int_\mathbf{z} p_{\bm\theta} (\mathbf{x}|\mathbf{z})  p (\mathbf{z}) d\mathbf{z}$.
	However it is intractable to compute the marginal log-likelihood without using simplifying assumption such as mean-field approximation \cite{blei2017variationalinference}.
	Therefore the following variational lower bound $-\mathcal{L}$ is used as a surrogate objective:
	\begin{multline*}
	-\mathcal{L}(\bm\theta, \bm\phi; \mathbf{x}) = -D_{KL}(q_{\bm\phi}(\mathbf{z}|\mathbf{x}) \Vert p (\mathbf{z}) ) \\
	+ \mathbb{E}_{q_{\bm\phi} (\mathbf{z}|\mathbf{x})} \left[ \log p_{\bm\theta} (\mathbf{x} | \mathbf{z}) \right],
	\end{multline*}
	where $q_{\bm\phi} (\mathbf{z}|\mathbf{x})$ is a variational approximation to the unknown $p_{\bm\theta} (\mathbf{z}|\mathbf{x})$, and $D_{KL}(q\Vert p)$ is the Kullback-Leibler (KL) divergence between $q$ and $p$.
	Maximizing the surrogate objective $-\mathcal{L}$ is proven to minimize $D_{KL} (q_{\bm\phi} (\mathbf{z}|\mathbf{x}) \Vert p_{\bm\theta} (\mathbf{z} | \mathbf{x}) )$,
	and it can also be seen as maximizing the expected data log-likelihood with respect to $q_{\bm\phi}$ while using $D_{KL} (q_{\bm\phi} (\mathbf{z}|\mathbf{x}) \Vert p_{\bm\theta} (\mathbf{z}) )$ as a regularization term.
	
	VAEs are successfully applied in modeling various data: including image \cite{pu2016imagevae,gulrajani2017pixelvae}, music \cite{roberts2018musicvae}, and text \cite{miao2016textvae,bowman2016textvae}.
	The VAE framework can also be extended to constructing conditional generative models \cite{sohn2015cvae} or learning from semi-supervised data \cite{kingma2014ssvae,xu2017ssvae}.
	
	\paragraph{VAEs for text pair modeling.}
	The most simple approach to modeling text pairs using the VAE framework is to consider two text sequences separately \cite{zhao2018arae,shen2018lvm}.
	That is, a generator is trained to reconstruct a single input sequence rather than integrating both sequences, and the two latent representations encoded from a variational posterior are given to a classifier network.
	When label information is not available, only the reconstruction objective is used for training.
	This means that the classifier parameters are not updated in the unsupervised setting, and thus the interaction between the variational posterior (or encoder) and the classifier could be restricted.
	
	\subsection{von Mises--Fisher Distribution}
	\label{subsec:vmf}
	Since the advent of deep generative models with variational inference, the typical choice for prior and variational posterior distribution has been the Gaussian, likely due to its well-studied properties and easiness of reparameterization.
	However it often leads a model to face the posterior collapse problem where a model ignores latent variables by pushing the KL divergence term to zero \cite{chen2017vlae,vandenoord2017vq}, 
	especially in text generation models where powerful decoders are used \cite{bowman2016textvae,yang2017dilated}.
	
	Various techniques are proposed to mitigate this problem:
	including KL cost annealing \cite{bowman2016textvae}, weakening decoders \cite{yang2017dilated}, skip connection \cite{dieng2019avoiding}, using different objectives \cite{alemi2018brokenelbo}, and using alternative distributions \cite{guu2018editing}.
	In this work, we take the last approach by utilizing a von Mises--Fisher (vMF) distribution.
	
	A vMF distribution is a probability distribution on the $(d-1)$-sphere, therefore samples are compared according to their directions, reminiscent of the cosine similarity.
	It has two parameters---mean direction $\bm\mu \in \mathbb{R}^{d}$ and concentration $\kappa \in \mathbb{R}$.
	As the KL divergence between $\text{vMF}(\bm\mu,\kappa)$ and the hyperspherical uniform distribution $\mathcal{U}(S^{d-1}) = \text{vMF}(\cdot, 0)$ only depends on $\kappa$, the KL divergence is a constant if the concentration parameter is fixed.
	Therefore when $\text{vMF}(\bm\mu,\kappa)$ with fixed $\kappa$ and $\text{vMF}(\cdot,0)$ are used as posterior and prior, the posterior collapse does not occur inherently.
	
	To the best of our knowledge, \citet{guu2018editing} were the first to use vMF as posterior and prior for VAEs, and \citet{xu2018spherical} empirically proved the effectiveness of vMF-VAE in natural language generation.
	\citet{davidson2018svae} generalized the vMF-VAE and proposed the reparameterization trick for vMF.
	We refer readers to Appendix \ref{app:vmf} for detailed description of vMF we used.
	
	\section{Proposed Framework}
	\label{sec:framework}
\begin{figure}
    	\centering
    	\includegraphics[width=0.7\linewidth, keepaspectratio]{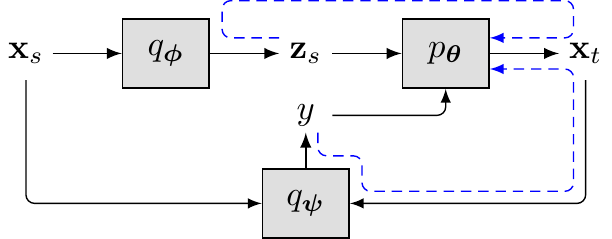}
    	\caption{The overview of the entire framework. Blue dashed lines indicate semantic constraints.}
    	\label{fig:model-overview}
	\end{figure}

	In this section, we describe the proposed framework in detail.
	We formally define the cross-sentence latent variable model (CS-LVM) and describe the optimization objectives. We also introduce semantic constraints to keep learned representations in a semantically plausible region.
	Fig. \ref{fig:model-overview} illustrates the entire framework.
	
	\begin{figure}
	\centering
	\begin{subfigure}{0.2\textwidth}
		\centering
		\includegraphics[height=4.25em, keepaspectratio]{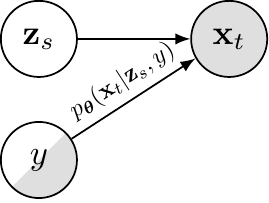}
		\caption{Generation model}
		\label{subfig:graphical-model-generation}
	\end{subfigure}%
	\begin{subfigure}{0.28\textwidth}
		\centering
		\includegraphics[height=4.25em, keepaspectratio]{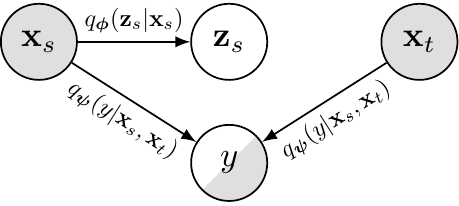}
		\caption{Recognition model}
		\label{subfig:graphical-model-recognition}
	\end{subfigure}
	\caption{Illustration of the graphical models. (a) the generative process of the output $\mathbf{x}_t$; (b) the approximate inference of $\mathbf{z}_s$ and the discriminative classifier for $y$.
	}
	\label{fig:graphical-model}
	\end{figure}
	
	\subsection{Cross-Sentence Latent Variable Model}
	\label{subsec:cslvm}
	Though the auto-encoding frameworks described in \S\ref{subsec:vae} have intriguing properties,
	it may hinder the possibility of training an encoder to extract rich features for text pair modeling,
	due to the fact that the generative modeling process is confined within a single sequence.
	Therefore the interaction between a generative model and a discriminative classifier is restricted,
	since the two sequences are separately modeled and the pair-wise information is only considered through the classifier network.

	Our proposed CS-LVM addresses this problem by cross-sentence generation of text given a text pair and its label.
	As the sentences in a pair are directly related within a generative model, the training objectives are defined in a more principled way than VAE-based semi-supervised text matching frameworks.
	Notably it also mimics the dataset construction process of some corpora: \emph{a worker generates a target text given a label and a source text} (e.g. \citealp{bowman2015snli,williams2018mnli}).
	
	Given a pair $(\mathbf{x}_1, \mathbf{x}_2)$, let $\mathbf{x}_s$, $\mathbf{x}_t \in \{\mathbf{x}_1, \mathbf{x}_2\}$ be a source and a target sequence respectively.
	Then we assume $\mathbf{x}_t$ is generated according to the following process (see Fig. \ref{subfig:graphical-model-generation}):
	\begin{enumerate}
		\item a latent variable $\mathbf{z}_s$ that contains the content of a source sequence is sampled from $p(\mathbf{z}_s)$,
		\item a variable $y$ that determines the relationship between a target and the source sequence is sampled from $p(y)$,
		\item $\mathbf{x}_t$ is generated from a conditional distribution $p_{\bm\theta}(\mathbf{x}_t|\mathbf{z}_s,y)$.
	\end{enumerate}
	In the above process, the class label $y$ is treated as a hidden variable in the unsupervised case and an observed variable in the supervised case.
	
	Accordingly, when the label information is available, the optimization objective for a generative model is the marginal log-likelihood of the observed variables $\mathbf{x}_t$ and $y$:
	\begin{multline}
	\label{eqn:true-objective}
	\log p_{\bm\theta} (\mathbf{x}_t, y)
	= \log \int p_{\bm\theta} (\mathbf{x}_t, \mathbf{z}_s, y) d\mathbf{z}_s
	\\= \log \int p_{\bm\theta} (\mathbf{x}_t | \mathbf{z}_s, y) p(\mathbf{z}_s) p(y) d\mathbf{z}_s.
	\end{multline}
	To address the intractability we instead optimize the lower bound of Eq. \ref{eqn:true-objective}:\footnote{See Appendix \ref{app:lb-derivation} for derivation of the lower bound.}
	\begin{multline}
	\label{eqn:elbo}
	\log p_{\bm\theta} (\mathbf{x}_t,y) \ge - D_{KL} ( q_{\bm\phi} (\mathbf{z}_s | \mathbf{x}_s) \Vert p (\mathbf{z}_s) ) \\
	+ \mathbb{E}_{q_{\bm\phi} (\mathbf{z}_s|\mathbf{x}_s)} [\log p_{\bm\theta} (\mathbf{x}_t | y, \mathbf{z}_s)] + \log p(y),
	\end{multline}
	where $q_{\bm\phi} (\mathbf{z}_s|\mathbf{x}_s)$ is a variational approximation of the posterior $p_{\bm\theta} (\mathbf{z}_s|\mathbf{x}_t, y)$.
	Though Eq. \ref{eqn:elbo} holds for any $q_{\bm\phi}$ having the same support with $p(\mathbf{z}_s)$, we choose this form of variational posterior from the following motivation:
	\emph{since $\mathbf{x}_s$ is related to $\mathbf{x}_t$ by the label information $y$, 
		$\mathbf{x}_s$ would have an influence on the space of $\mathbf{z}_s$ in a similar way to $(\mathbf{x}_t, y)$.}
	Due to this particular choice of $q_{\bm\phi}$,
	$\mathbf{z}_s$ depends only on $\mathbf{x}_s$ and is independent of the label information possibly permeated in $\mathbf{x}_t$.
	In other words, this design induces $q_{\bm\phi}$ to extract the features needed for controlling the semantics only from $\mathbf{x}_s$, while preventing $q_{\bm\phi}$ from encoding other biases.
	
	To extend the objective to the unsupervised setup, we marginalize out $y$ from Eq. \ref{eqn:elbo} using a classifier distribution.
	We will provide more detailed explanation of the optimization objectives in \S\ref{subsec:optimization}.
	
	\subsection{Architecture}
	\label{subsec:architecture}
	Now we describe the architectures we used for constructing CS-LVM.
	We first encode a source sequence into a fixed-length representation using a recurrent neural network (RNN): $g^{enc}(\mathbf{x}_s) = \mathbf{m}_s$.
	From $\mathbf{m}_s$ we obtain a variational approximate distribution $q_{\bm\phi} (\mathbf{z}_s|\mathbf{x}_s) = g^{code} (\mathbf{m}_s)$
	and sample a latent representation $\mathbf{z}_s \sim q_{\bm\phi} (\mathbf{z}_s|\mathbf{x}_s)$.
	In our experiments, a long short-term memory (LSTM) recurrent network and a feed-forward network are used as $g^{enc}$ and $g^{code}$ respectively.
	From the fact that the mean direction parameter $\bm\mu_s$ of $\text{vMF}(\bm\mu_s,\kappa)$ should be a unit vector, $g^{code}$ additionally normalizes the output of the feed-forward network to be $\lVert g^{code}(\mathbf{m}_s) \rVert_2 = 1$.
	
	Then we generate the target sequence $\mathbf{x}_t$ from $\mathbf{z}_s$ and $y$.
	Similarly to the encoder network, we use an LSTM for a decoder, thus the distribution is factorized as follows:
	\begin{equation}
	\label{eqn:factorize}
	p_{\bm\theta} (\mathbf{x}_t | y,\mathbf{z}_s) = \prod_{i=1}^{N_{\mathbf{x}_t} + 1} p_{\bm\theta} (w_{t,i}|w_{t,<i},y,\mathbf{z}_s),
	\end{equation}
	where $\mathbf{x}_t=(x_{t,1}, \ldots, x_{t,N_{\mathbf{x}_t}})$ and $w_{t,0}=\texttt{<s>}$, $w_{t,N_{\mathbf{x}_t} + 1}=\texttt{</s>}$ are special tokens indicating the start and the end of a sequence.
	
	We project the word index $w_{t,i}$ and label index $y$ into embedding spaces to obtain the word embedding $\mathbf{w}_{t,i}$ and label embedding $\mathbf{y}$.
	Then to construct an input for $i$-th time step, $\mathbf{v}_t$, we concatenate the $i$-th target word embedding $\mathbf{w}_{t,i}$, the label embedding $\mathbf{y}$, and the latent representation $\mathbf{z}_s$ altogether:
	
	\begin{equation*}
	\mathbf{v}_i = [\mathbf{w}_{t,i}; \mathbf{y}; \mathbf{z}_s ].
	\end{equation*}
	Thus $p_{\bm\theta} (w_{t,i}|w_{t,<i},\mathbf{z}_s,y)$ is computed from $i$-th state $\mathbf{s}_i$ of the decoder RNN:
	\begin{equation*}
	p_{\bm\theta} (w_{t,i}|w_{t,<i},y,\mathbf{z}_s) = \text{softmax}(g^{out} (\mathbf{s}_i))
	\end{equation*}
	\begin{equation*}
	\mathbf{s}_i = g^{dec}_i (\mathbf{v}_i, \mathbf{s}_{i-1}),
	\end{equation*}
	where $g^{out}$ is a feed-forward network and $g^{dec}_i$ is the state transition function of the decoder LSTM at $i$-th time step.
	
	For a discriminative classifier network we follow the siamese architecture, as mentioned in \S\ref{sec:intro}.
	$\mathbf{x}_s$ and $\mathbf{x}_t$ are fed to a shared LSTM network $f^{enc}$ to obtain sentence vectors $\mathbf{h}_1 = f^{enc} (\mathbf{x}_s)$ and $\mathbf{h}_2=f^{enc} (\mathbf{x}_t)$.
	Then $\mathbf{h}_1$ and $\mathbf{h}_2$ are combined by the function $f^{fuse}$ to form a single fused vector,
	and the fused representation is given to a feed-forward network $f^{disc}$ to infer the relationship:
	\begin{equation*}
	q_{\bm\psi} (y|\mathbf{x}_1,\mathbf{x}_2) = \text{softmax}(f^{disc} (f^{fuse}(\mathbf{h}_1, \mathbf{h}_2))).
	\end{equation*}
	
	To learn from data more efficiently and to reduce the number of trainable parameters,
	we tie the weights for two encoders---for the generative model and the discriminative classifier; i.e. $g^{enc}=f^{enc}$.
	This mitigates the problem that only source sequences are used for training $g^{enc}$
	and enhances the interaction between the generative model and the classifier.
	We will see from experiments that tying encoder weights improves performance and stabilizes optimization (\S\ref{subsec:ablation}).
	
	Also note that the functions $g^{\square}$ are only used in training, and the model has the same test-time computational complexity with typical classification models.
	
	\subsection{Optimization}
	\label{subsec:optimization}
	
	\begin{algorithm}[t]
		\small
		\begin{tabular}{l@{ }l}
			\textbf{Input:} & Labeled dataset $\mathcal{X}_l$,
				Unlabeled dataset $\mathcal{X}_u$, \\
			& Model parameters $\bm\theta, \bm\phi, \bm\psi$
		\end{tabular}
		\begin{algorithmic}[1]
			\Procedure{Train}{$\mathcal{X}_l, \mathcal{X}_u, \bm\theta, \bm\phi, \bm\psi$}
			\Repeat
			\State{Sample $(\mathbf{x}_{l,s}, \mathbf{x}_{l,t}, y_l) \sim \mathcal{X}_l$} \label{lst:alg:base-start}
			\State{Sample $(\mathbf{x}_{u,s}, \mathbf{x}_{u,t}) \sim \mathcal{X}_u$}
			\State{Compute $\mathcal{L}_l(\bm\theta,\bm\phi,\bm\psi;\mathbf{x}_{l,s},\mathbf{x}_{l,t},y_l)$ by (\ref{eqn:supervised-objective})}
			\State{Compute $\mathcal{L}_u(\bm\theta,\bm\phi,\bm\psi;\mathbf{x}_{u,s},\mathbf{x}_{u,t})$ by (\ref{eqn:unsupervised-objective})}
			\State{Update $\bm\theta, \bm\phi, \bm\psi$ by gradient descent on $\mathcal{L}_l + \mathcal{L}_u$} \label{lst:alg:base-end}
			\Until stop criterion is met
			\EndProcedure
			
			\Procedure{FineTune}{$\mathcal{X}_l, \mathcal{X}_u, \bm\theta, \bm\phi, \bm\psi$}
			\Repeat
			\State{Update $\bm\theta, \bm\phi, \bm\psi$ following line \ref{lst:alg:base-start}--\ref{lst:alg:base-end}}
			\parState{Update $\bm\theta$ by gradient descent on (\ref{eqn:constraint-y}--\ref{eqn:constraint-multi})}
			\Until stop criterion is met
			\EndProcedure
		\end{algorithmic}
		\caption{Training procedure of CS-LVM.}
		\label{alg:overall}
	\end{algorithm}

	In this subsection we describe how the entire model is optimized.
	We first define optimization objectives for supervised and unsupervised training,
	and then introduce constraints to regularize the model to generate sequences with intended semantic characteristics.
	The entire optimization procedure is summarized in Algorithm \ref{alg:overall}.
	
	\subsubsection{Supervised Objective}
	\label{subsubsec:supervised-training}
	In the supervised setting, a data sample is assumed to contain label information: $(\mathbf{x}_1, \mathbf{x}_2, y) \in \mathcal{X}_l$.
	Without loss of generality let us assume $(\mathbf{x}_s, \mathbf{x}_t) = (\mathbf{x}_1, \mathbf{x}_2)$.\footnote{%
		The relationship between a source and a target may either be unidirectional, bidirectional, or reflexive, depending on the characteristics of a task.
		For some experiments we additionally used swapped data examples, $(\mathbf{x}_s, \mathbf{x}_t)=(\mathbf{x}_2, \mathbf{x}_1)$, for training.
		We explain more on this in \S\ref{sec:experiments}.
	}
	Since $y$ is an observed variable in this case, we can directly use Eq. \ref{eqn:elbo} in training.
	From Eqs. \ref{eqn:elbo} and \ref{eqn:factorize}, the objective for the generative model is defined by:\footnote{%
		Note that we define all objectives $\mathcal{L}$, $\mathcal{R}$ as \emph{minimization
			objectives} to avoid confusion.
	}
	\begin{multline}
	\label{eqn:labeled-gen-objective}
	-\mathcal{L}_l^{gen} (\bm\theta, \bm\phi; \mathbf{x}_s, \mathbf{x}_t, y)
	= \log p_{\bm\theta} (\mathbf{x}_t|y,\mathbf{z}_s) \\
	+ \log p(y)
	- D_{KL} ( q_{\bm\phi} (\mathbf{z}_s | \mathbf{x}_s) \Vert p (\mathbf{z}_s) ),
	\end{multline}
	where $\mathbf{z}_s \sim q_{\bm\phi} (\mathbf{z}_s | \mathbf{x}_s)$ and $p(y)$, $p (\mathbf{z}_s)$ are prior distributions of $y$, $\mathbf{z}_s$.
	Considering that we assume $p (y)$ to be a fixed uniform distribution of labels, the $\log p(y)$ term can be ignored in training: $\lVert \nabla_{\bm\theta,\bm\phi} \log p(y)\rVert_2 = 0$.
	
	For training, the typical teacher forcing method is used; i.e. ground-truth words are used as input words.
	We use $\text{vMF}(g^{code}(\mathbf{m}_s), \kappa)$ ($\kappa$: hyperparameter) for the variational posterior $q_{\bm\phi}(\mathbf{z}_s|\mathbf{x}_s)$ and $\text{vMF}(\cdot, 0)$ for the prior $p(\mathbf{z}_s)$.
	
	The discriminator objective is defined as a conventional maximum likelihood:
	\begin{equation}
	-\mathcal{L}_l^{disc} (\bm\psi; \mathbf{x}_s, \mathbf{x}_t, y) = \log q_{\bm\psi} (y|\mathbf{x}_s,\mathbf{x}_t).
	\end{equation}
	
	Finally, the two objectives are combined to construct the objective for supervised training:
	\begin{equation}
	\label{eqn:supervised-objective}
	\mathcal{L}_l(\bm\theta,\bm\phi,\bm\psi;\mathbf{x}_s,\mathbf{x}_t,y) = \mathcal{L}_l^{gen} + \lambda \mathcal{L}_l^{disc},
	\end{equation}
	where $\lambda$ is a hyperparameter.
	
	\subsubsection{Unsupervised Objective}
	\label{subsubsec:unsupervised-training}
	In this case, the model does not have an access to label information; a data point is represented by $(\mathbf{x}_s, \mathbf{x}_t) \in \mathcal{X}_u$ and thus $y$ is a hidden variable.
	To facilitate the unsupervised training, we marginalize $y$ out as below and derive the lower bound:
	\begin{multline}
	\label{eqn:unsupervised-marginalization}
	\log p_{\bm\theta} (\mathbf{x}_t) = \log \sum_{y} \int p_{\bm\theta} (\mathbf{x}_t, \mathbf{z}_s, y)  d\mathbf{z}_s \\
	\ge \mathbb{E}_{q_{\bm\phi,\bm\psi}(y,\mathbf{z}_s|\mathbf{x}_s,\mathbf{x}_t)}
	\left[
	\log
	\frac{p_{\bm\theta}(\mathbf{x}_t, \mathbf{z}_s, y)}
	{q_{\bm\phi,\bm\psi}(y,\mathbf{z}_s|\mathbf{x}_s,\mathbf{x}_t)}
	\right].
	\end{multline}
	And from the assumption presented in the graphical model (Fig. \ref{subfig:graphical-model-recognition}),
	\begin{equation}
	\label{eqn:unsupervised-factorization} q_{\bm\phi,\bm\psi}(y,\mathbf{z}_s|\mathbf{x}_s,\mathbf{x}_t)
	= q_{\bm\phi}(\mathbf{z}_s|\mathbf{x}_s)
	q_{\bm\psi}(y|\mathbf{x}_s,\mathbf{x}_t).
	\end{equation}
	Finally we obtain the following lower bound for $\log p_{\bm\theta}(\mathbf{x}_t)$ from Eqs. \ref{eqn:unsupervised-marginalization} and \ref{eqn:unsupervised-factorization}:\footnote{See Appendix \ref{app:lb-derivation} for details.}
	\begin{multline}
	\label{eqn:unsupervised-objective}
	\mathcal{L}_u(\bm\theta,\bm\phi,\bm\psi;\mathbf{x}_s,\mathbf{x}_t) = - \mathcal{H}(q_{\bm\psi}(y|\mathbf{x}_s,\mathbf{x}_t)) \\
	+ \mathbb{E}_{q_{\bm\psi}(y|\mathbf{x}_s,\mathbf{x}_t)} \left[ \mathcal{L}_l^{gen}(\bm\theta,\bm\phi;\mathbf{x}_s,\mathbf{x}_t,y) \right].
	\end{multline}
	Here the second expectation term can be computed either by enumeration or sampling,
	and we used the former as the datasets we used have relatively small label sets (2 or 3) and it is known to yield better results than sampling \cite{xu2017ssvae}.
	We will compare the two methods in \S\ref{subsec:ablation}.
	
	To sum up, at every training iteration, given a labeled and unlabeled data sample $(\mathbf{x}_{l,s}, \mathbf{x}_{l,t}, y_l)$, $(\mathbf{x}_{u,s}, \mathbf{x}_{u,t})$, we optimize the following objective.
	\begin{multline}
	    \mathcal{L} = \mathcal{L}_l(\bm\theta,\bm\phi,\bm\psi;\mathbf{x}_{l,s},\mathbf{x}_{l,t},y_l)
	    \\ + \mathcal{L}_u(\bm\theta,\bm\phi,\bm\psi;\mathbf{x}_{u,s},\mathbf{x}_{u,t})
	\end{multline}
	
	\subsubsection{Fine-Tuning with Semantic Constraints}
	\label{subsubsec:semantic-constraints}
	Since the generator is trained via maximum likelihood training which considers all words in a sentence equivalently,
	the label information may not be reflected enough in generation owing to high-frequency words.
	For example in natural language inference, the word occurrences of the following three hypothesis sentences highly overlap, but they should have different relation with the premise.\footnote{Examples are taken from the SNLI development set, pair ID \texttt{4904199439.jpg\#\{2r1e,2r1n,2r1c\}}.}
	\begin{addmargin}[0.5em]{0.25em}
		\normalsize
		\smallskip
		\textbf{P}: \textit{A man is cutting metal with a tool .} \\
		\textbf{H1}: \textit{A man is cutting metal .} \\
		\textbf{H2}: \textit{A man is cutting metal with the wrong tool .} \\
		\textbf{H3}: \textit{A man is cutting metal with his mind .} 
		\smallskip
	\end{addmargin}
	Thus for some data points, the strategy that only predicts words that overlap across hypotheses could receive a fairly high score, which might weaken the integration of $y$ into the generator.
	To mitigate this, we fine-tune the trained generator using the following semantic constraint:
	\begin{equation}
	\label{eqn:constraint-y}
	-\mathcal{R}^{y} (\bm\theta; \mathbf{x}_s,\mathbf{x}_t)
	= \log q_{\bm\psi} (\tilde{y} | \mathbf{x}_s, \widetilde{\mathbf{x}}_t),
	\end{equation}
	where $\tilde{y} \sim p(y)$, $\mathbf{z}_s \sim q_{\bm\phi} (\mathbf{z}_s | \mathbf{x}_s)$, and $	\widetilde{\mathbf{x}}_t = \argmax_{\mathbf{x}_t} p_{\bm\theta} (\mathbf{x}_t | \tilde{y}, \mathbf{z}_s)$.
	This constraint enforces the sequence $\widetilde{\mathbf{x}}_t$ generated by conditioning on $\tilde{y}$ and $\mathbf{z}_s$ to actually have the relationship $\tilde{y}$ with $\mathbf{x}_s$.
	
	We also introduce a constraint on $\mathbf{z}$ that keeps the distributions of $\widetilde{\mathbf{z}}_t$ (the latent content variable obtained by encoding the generated sequence $\widetilde{\mathbf{x}}_t$) and $\mathbf{z}_s$ close:
	\begin{equation}
	\label{eqn:constraint-z}
	-\mathcal{R}^{\mathbf{z}} (\bm\theta; \mathbf{x}_s,\mathbf{x}_t)
	= \log q_{\bm\phi} (\mathbf{z}_t = \widetilde{\mathbf{z}}_t | {\mathbf{x}}_t),
	\end{equation}
	where $\widetilde{\mathbf{z}}_t \sim q_{\bm\phi} (\widetilde{\mathbf{z}}_t | \widetilde{\mathbf{x}}_t)$.
	In other words, it pushes the generated sequence $\widetilde{\mathbf{x}}_t$ to be in a similar semantic space with the ground-truth target sequence $\mathbf{x}_t$.
	Consequently, it can help alleviate the generator collapse problem where a generator produces only a handful of simple neutral patterns independent of the input sequence, by relating $\widetilde{\mathbf{z}}_t$ to $\mathbf{z}_t$.\footnote{%
		The basic assumption behind this constraint is that a source and a target sequence are associated in a certain aspect,
		and it generally holds in most of the available pair classification datasets e.g. SNLI, SICK, SciTail, QQP, MRPC.
	}
	
	From similar motivation, we also add an additional constraint that encourages the generated sentences originating from different source sentences to be dissimilar.
	To reflect this, we define the following minibatch-level constraint that penalizes the mean direction vectors encoded from the generated sentences for being too close:
	\begin{equation}
	\label{eqn:constraint-mu}
	\begin{gathered}
	-\mathcal{R}^{\bm\mu}(\bm\theta; \mathcal{B}) = \mathbb{E}_{\mathcal{B}} [d( {\bm\mu_t}^{(i)}, \bar{\bm\mu}_t)],
	\end{gathered}
	\end{equation}
	where
	we denote values related to $i$-th sample of a minibatch $\mathcal{B}$ using superscript: $\square^{(i)}$.
	In the above, $\bm\mu_t^{(i)} = g^{code} (g^{enc}(\widetilde{\mathbf{x}}_t^{(i)}))$, 
	$\bar{\bm\mu}_t = \sum_{i=1}^{|\mathcal{B}|} \bm\mu_t^{(i)}/|\mathcal{B}|$,
	and $d(\cdot, \cdot)$ is a distance measure between vectors.
	The mean direction vector $\bm\mu$ of $\text{vMF}(\bm\mu,\kappa)$ is on a unit hypersphere, so we use the cosine distance: $d(\bm\mu_1,\bm\mu_2) = 1 - \langle \bm\mu_1, \bm\mu_2 \rangle$.
	
	As the sequence generation process is not differentiable, the gradients from the semantic constraints cannot propagate to the generator parameters.
	To relax the discreteness, we use the Gumbel-Softmax reparameterization \cite{jang2017gumbel,maddison2017concrete}.
	Using the Gumbel-Softmax trick, we obtain a continuous probability vector that approximates a sample from the categorical distribution of words at each step,
	and use the probability vector to compute the expected word embedding for the subsequent step.
	
	When multiple constraints are used, they are combined using the homoscedastic uncertainty weighting \cite{kendall2018uncertainty}:\footnote{%
		Though the weighting scheme is originally derived from the case of a Gaussian likelihood,
		\citet{kendall2018uncertainty,xiong2018dcn,hu2018rmreader} successfully applied it in weighting various losses e.g. cross-entropy loss, $L_1$ loss, and reinforcement learning objectives.
	}
	\begin{multline}
	\label{eqn:constraint-multi}
	\mathcal{R} = \frac{1}{\sigma_1^2}\mathcal{R}^{y}
	 + \frac{1}{\sigma_2^2}\mathcal{R}^{\mathbf{z}}
	 + \frac{1}{\sigma_3^2}\mathcal{R}^{\bm\mu}
	 \\ + \log \sigma_1 + \log \sigma_2 + \log \sigma_3,
	\end{multline}
	where $\sigma_1, \sigma_2, \sigma_3$ are trainable scalar parameters.
	Also note that all constraints are \emph{unsupervised}, where label information is not required.
	
	\section{Experiments}
	\label{sec:experiments}
	
	\begin{table}[tb]
		\centering
		\small
		\begin{tabular}{l r r r}
			\hline
			\textbf{Model} & \textbf{28k} & \textbf{59k} & \textbf{120k} \\
			\hline
			LSTM$^\text{(a)}$ & 57.9 & 62.5 & 65.9  \\
			CNN$^\text{(b)}$ & 58.7 & 62.7 & 65.6 \\
			LSTM-AE$^\text{(a)}$ & 59.9 & 64.6 & 68.5 \\
			LSTM-ADAE$^\text{(a)}$ & 62.5 & 66.8 & 70.9  \\
			DeConv-AE$^\text{(b)}$ & 62.1 & 65.5 & 68.7 \\
			LSTM-VAE$^\text{(b)}$ & 64.7 & 67.5 & 71.1  \\
			DeConv-VAE$^\text{(b)}$ & 67.2 & 69.3 & 72.2 \\
			LSTM-vMF-VAE (ours) & 65.6 & 68.7 & 71.1 \\
			\hline
			CS-LVM (ours) & 68.4 & 73.5 & 76.9 \\
			\quad $+ \mathcal{R}^{y}$ & \textbf{70.0} & \textbf{74.5} & 77.4 \\
			\quad $+ \mathcal{R}^{\mathbf{z}}$ & 69.2 & 73.9 & 77.6 \\
			\quad $+ \mathcal{R}^{\bm\mu}$ & 69.1 & 74.0 & \textbf{77.6} \\
			\quad $+ \mathcal{R}^{y}, \mathcal{R}^{\mathbf{z}}, \mathcal{R}^{\bm\mu}$ & 69.6 & 74.1 & 77.4 \\
			\hline
		\end{tabular}
		\caption{
			Semi-supervised classification results on the SNLI dataset.
			(a) \citet{zhao2018arae}; (b) \citet{shen2018lvm}.
		}
		\label{table:snli}
	\end{table}
	
	\begin{table}[tb]
		\centering
		\small
		\begin{tabular}{l r r r r}
			\hline
			\textbf{Model} & \textbf{1k} & \textbf{5k} & \textbf{10k} & \textbf{25k} \\
			\hline
			CNN$^\text{(a)}$ & 56.3 & 59.2 & 63.8 & 68.9 \\
			LSTM-AE$^\text{(a)}$ & 59.3 & 63.8 & 67.2 & 70.9 \\
			DeConv-AE$^\text{(a)}$ & 60.2 & 65.1 & 67.7 & 71.6 \\
			LSTM-VAE$^\text{(a)}$ & 62.9 & 67.6 & 69.0 & 72.4 \\
			DeConv-VAE$^\text{(a)}$ & 65.1 & 69.4 & 70.5 & 73.7 \\
			LSTM-vMF-VAE (ours) & 65.0 & 69.9 & 72.1 & 74.9 \\
			\hline
			CS-LVM (ours) & \bf 66.5 & 71.1 & 74.6 & 76.9 \\
			\quad $+ \mathcal{R}^{y}$ & 66.4 & 70.8 & 74.5 & 77.5 \\
			\quad $+ \mathcal{R}^{\mathbf{z}}$ & \bf 66.5 & \bf 71.3 & 74.8 & 77.1 \\
			\quad $+ \mathcal{R}^{\bm\mu}$ & 66.4 & 71.2 & \textbf{74.9} & 77.4 \\
			\quad $+ \mathcal{R}^{y}, \mathcal{R}^{\mathbf{z}}, \mathcal{R}^{\bm\mu}$ & 66.3 & \bf 71.3 & 74.7 & \textbf{77.6} \\
			\hline
		\end{tabular}
		\caption{
			Semi-supervised classification results on the Quora Question Pairs dataset.
			(a) \citet{shen2018lvm}.
		}
		\label{table:quora}
	\end{table}
	
	We evaluate the proposed model on two semi-supervised tasks: \emph{natural language inference} and \emph{paraphrase identification}.
	We also implement a strong baseline that has a similar architecture to LSTM-VAE \cite{shen2018lvm} but uses vMF distribution for prior and posterior, named LSTM-vMF-VAE.
	To further explore the proposed model, we conduct extensive qualitative analyses. %
	For detailed settings and hyperparameters, please refer to Appendix \ref{app:implementation}.
\
	\subsection{Natural Language Inference}
	Natural language inference (NLI) is a task of predicting the relationship given a premise and a hypothesis sentence.
	We use Stanford Natural Language Inference (SNLI, \citealp{bowman2015snli}) dataset for experiments.
	It consists of roughly 570k premise-hypothesis pairs, and each pair has one of the following labels: \emph{entailment}, \emph{neutral}, and \emph{contradiction}.
	Considering the asymmetry in some label classes and for conformance with the dataset generation process, we use premise and hypothesis sentence as source and target respectively:
	$(\mathbf{x}_s, \mathbf{x}_t)=(\mathbf{x}_{pre}, \mathbf{x}_{hyp})$.
	
	Following the work of \citet{zhao2018arae,shen2018lvm}, we consider scenarios where 28k, 59k, and 120k labeled data samples are available.
	Also, for fair comparison with the prior work, we set the size of a word vocabulary set to 20,000 and do not utilize pre-trained word embeddings such as GloVe \cite{pennington2014glove}.
	
	To combine the representations of a premise and a hypothesis and to construct an input to $f^{disc}$, we use the following heuristic-based fusion proposed by \citet{mou2016snli}:
	\begin{multline}
	f^{fuse}(\mathbf{h}_{pre}, \mathbf{h}_{hyp}) \\ = \left[
	\mathbf{h}_{pre}; \mathbf{h}_{hyp};
	|\mathbf{h}_{pre} - \mathbf{h}_{hyp}|;
	\mathbf{h}_{pre} \odot \mathbf{h}_{hyp}
	\right],
	\end{multline}
	where $[\mathbf{a};\mathbf{b}]$ indicates concatenation of vectors $\mathbf{a}$, $\mathbf{b}$ and $\odot$ is the element-wise product.
	
	Table \ref{table:snli} summarizes the result of experiments.
	We can clearly see that the proposed CS-LVM architecture substantially outperforms other models based on auto-encoding.
	Also, the semantic constraints brought additional boost in performance, achieving the new state of the art in semi-supervised classification of the SNLI dataset.
	
	When all training data are used as labeled data ($\approx$ 550k), CS-LVM also improves performance by achieving accuracy of 82.8\%, compared to the supervised LSTM (81.5\%), LSTM-AE (81.6\%), LSTM-VAE (80.8\%), DeConv-VAE (80.9\%).%
	\subsection{Paraphrase Identification}
	Paraphrase identification (PI) is a task whose objective is to infer whether two sentences have the same semantics.
	We use the Quora Question Pairs dataset (QQP, \citealp{wang2017bilateral}) for experiments.
	QQP consists of over 400k sentence pairs each of which has label information indicating whether the sentences in a pair paraphrase each other or not.
	We experiment for the cases where the number of labeled data is 1k, 5k, 10k, and 25k, and set the vocabulary size to 10,000, following \citet{shen2018lvm}.
	Unlike auto-encoding--based models that treat sentences in a pair equivalently, the CS-LVM processes them asymmetrically for its cross-sentence generating property.
	This property is useful when some relationships are asymmetric (e.g. NLI%
	),
	however the paraphrase relationship is bidirectional, so that we also use swapped text pairs in training.
	To fuse sentence representations, the following symmetric function is used, as in \citet{ji2013discriminative}:
	\begin{equation}
	f^{fuse} (\mathbf{h}_1, \mathbf{h}_2) = [\mathbf{h}_1 + \mathbf{h}_2; |\mathbf{h}_1 - \mathbf{h}_2|].
	\end{equation}
	
	The result of experiments on QQP is summarized in Table \ref{table:quora}.
	Again, the proposed CS-LVM consistently outperforms other supervised and semi-supervised models by a large margin,
	setting the new state-of-the-art result on the QQP dataset with the semi-supervised setting.
	
	\subsection{Ablation Study}
	\label{subsec:ablation}
	\begin{table}[tb]
		\centering
		\small
		\begin{tabular}{l r r r}
			\hline
			\textbf{Model} & \textbf{28k} & \textbf{59k} & \textbf{120k} \\
			\hline
			CS-LVM & 68.4 & 73.5 & 76.9 \\
			\quad \textit{(i) without CS} & 65.6 & 68.7 & 71.1 \\
			\quad \textit{(ii) Gaussian} & 66.9 & 72.0 & 74.9 \\
			\quad \textit{(iii) sampling} & 68.0 & 72.9 & 76.5 \\
			\quad \textit{(iv) $f^{enc}\neq g^{enc}$} & 63.3 & 69.1 & 74.7 \\
			\hline
		\end{tabular}
		\caption{Ablation study results.}
		\label{table:ablation}
	\end{table}
	To assess the effect of each element, we experiment with model variants where some of the components are removed.
	Specifically, we conduct an ablation study for the following variants:
	(i) without cross-sentence generation (i.e. auto-encoding setup),
	(ii) replacing the vMF distribution with Gaussian,
	(iii) computing the expectation term of Eq. \ref{eqn:unsupervised-objective} by sampling,
	and (iv) without encoder weight sharing (i.e. $f^{enc} \neq g^{enc}$).
	SNLI dataset is used for the model ablation experiments, and trained models are not fine-tuned in order to focus only on the efficacy of each model component.
	
	Results of ablation study are presented in Table \ref{table:ablation}.
	As expected, the cross-sentence generation is the most critical factor for the performance, except for the 28k setting where the encoder weight tying brought the biggest gain.
	In 59k and 120k settings, all other variants that maintain the cross-generating property outperform the VAE-based models (see \textit{(ii)}, \textit{(iii)}, \textit{(iv)}).
	
	Replacing a vMF with a Gaussian does not severely harm the accuracy, however it requires the additional process of finding a KL cost annealing rate.
	When sampling is used instead of enumeration for computing Eq. \ref{eqn:unsupervised-objective}, about 1.2x speedup is observed in exchange for slight performance degradation,
	and thus sampling could be a good option in the case that the number of label classes is large.
	
	Finally, as mentioned in \S\ref{subsec:architecture}, variants whose encoder weights are untied do not work well.
	We conjecture this is because $g^{enc}$ receives the error signal only from a source sentence and could not fully benefit from both sentences.
	The fact that the performance degradation is larger when the number of labeled data is small also agrees with our hypothesis, since unlabeled data affect the classifier encoder only by the entropy term when encoder weights are not shared.
	
	\subsection{Generated Sentences}
	\begin{table*}
		\centering
		\small
		\begin{tabular}{p{0.26\textwidth}|p{0.2\textwidth}|p{0.2\textwidth}|p{0.2\textwidth}}
			\hline
			\textbf{Input} & \textbf{Entailment} & \textbf{Neutral} & \textbf{Contradiction} \\
			\hline
			two girls play with bubbles near a boat dock .
				& two girls are outside .
				& the girls are friends .
				& two girls are swimming in the ocean . \\
			\hline
			a classroom full of men, with the teacher up front .
				& a group of boys are indoors .
				& the teacher is teaching the students .
				& the students are at home sleeping . \\
			\hline
			a dune buggy traveling on sand .
				& the vehicle is moving .
				& the vehicle is red .
				& a man is riding a bike . \\
			\hline
		\end{tabular}
		\caption{Selected samples generated from the model trained on the SNLI dataset.}
		\label{table:samples}
	\end{table*}
	\begin{table}
		\centering
		\small
		\begin{tabular}{l r r r}
			\hline
			\bf Dataset & \bf Acc. & \bf \textit{distinct-1} & \bf \textit{distinct-2} \\
			\hline
			CS-LVM & 76.5 & .0128 & .0441 \\
			\quad $+\mathcal{R}^y$ & 81.9 & .0135 & .0479 \\
			\quad $+\mathcal{R}^{\mathbf{z}}$ & 79.0 & .0140 & .0492 \\
			\quad $+\mathcal{R}^{\bm\mu}$ & 77.5  & .0141 & .0488 \\ %
			\hline
		\end{tabular}
		\caption{
			Results of evaluation of generated artificial datasets.
			\textit{distinct-1} and \textit{distinct-2} compute the ratio of the number of unique unigrams or bigrams to that of the total generated tokens \cite{li2016diversity}.
		}
		\label{table:eval-gen-datasets}
	\end{table}
	We give examples of generated sentences, to validate that the proposed model learns to generate text having desired properties.
	From Table \ref{table:samples}, we can see that sentences generated from the identical input sentence properly reflect the label information given.
	More generated examples are presented in Appendix \ref{app:generated-examples}.
	
	Further, to quantitatively measure the quality of generated sentences, we construct artificial datasets,
	where each premise and label in the SNLI development set is used as input to our trained generator and generated hypotheses are collected.
	Then we prepare a LSTM classifier that is trained on the original SNLI dataset as a surrogate for the ideal classifier, and use it for measuring the quality of generated datasets.\footnote{The accuracy of the trained classifier on the original development set is 81.7\%.}
	We also compute the diversity of the generated hypotheses using the metrics proposed by \citet{li2016diversity}, to verify the effect of diversity-promoting semantic constraints.
	
	Results of the evaluation on the artificial datasets are presented in Table \ref{table:eval-gen-datasets}.
	The classifier trained on the original dataset predicts the generated data fairly well, from which we verify that the generated sentences contain desired semantics.
	Also, as expected, fine-tuning with $\mathcal{R}^y$ increases the classification accuracy by a large margin, while $\mathcal{R}^{\mathbf{z}}$ and $\mathcal{R}^{\bm\mu}$ enhance diversity.
			
	\section{Related Work}
	\label{sec:related-work}
	\paragraph{Semi-supervised learning for text classification.}
	Using unlabeled data for text classification is an important subject and there exists much previous research \cite[][to name but a few]{zhu2003semisupervisedgaussian,nigam2006semi,zhu2008semisupervised}.
	Notably, the work of \citet{xu2017ssvae} applies the semi-supervised VAE \cite{kingma2014ssvae}  to the single-sentence text classification problem.
	\citet{zhao2018arae,shen2018lvm} present VAE models for the semi-supervised text sequence matching, while their models have drawbacks as mentioned in \S\ref{sec:framework}.
	
	When the use of external corpora is allowed, the performance can further be increased.
	\citet{dai2015semi,ramachandran2017unsupervised} train an encoder-decoder network on large corpora and fine-tune the learned encoder on a specific task.
	Recently, there have been remarkable improvements in pre-trained language representations \cite{peters2018elmo,radford2018gpt,devlin2018bert},
	where language models trained on extremely large data brought a huge performance boost.
	These methods are orthogonal to our work, and additional enhancements are expected when they are used together with our model.
	
	\paragraph{Cross-sentence generating LVMs.}
	There exists some prior work on cross-sentence generating LVMs.
	\citet{shen2017crossalignment} introduce a similar data generation assumption to ours and apply the idea to unaligned style transfer and natural language generation.
	\citet{zhang2016vnmt,serban2017vhred} use latent variable models for machine translation and dialogue generation.
	\citet{kang2018adventure} propose a data augmentation framework for natural language inference that generates a sentence, however unlabeled data are not considered in their work.
	\citet{deudon2018semantic} build a sentence-reformulating deep generative model whose objective is to measure the semantic similarity between a sentence pair.
	However their work cannot be applied to a multi-class classification problem, and the generative objective is only used in pre-training, not considering the joint optimization of the generative and the discriminative objective. 
	To the best of our knowledge, our work is the first work on introducing the concept of cross-sentence generating LVM to the semi-supervised text matching problem.

	\section{Conclusion}
	\label{sec:conclusion}
	In this work, we proposed a cross-sentence latent variable model (CS-LVM) for semi-supervised text sequence matching.
	Given a pair of text sequences and the corresponding label, it uses one of the sequences and the label as input and generates the other sequence.
	Due to the use of cross-sentence generation, the generative model and the discriminative classifier interacts more strongly, and from experiments we empirically proved that the CS-LVM outperforms other models by a large margin.
	We also defined multiple semantic constraints to further regularize the model,
	and observed that fine-tuning with them gives additional increase in performance.
	
 	For future work, we plan to focus on generating more realistic text and use the generated text in other tasks e.g. data augmentation, addressing adversarial attack.
 	Although the current model makes fairly plausible sentences, it tends to prefer relatively short and \emph{safe} sentences, as the main goal of the training is to accurately predict the relationship between sentences.
 	We expect the model could perform more natural generation via applying recent advancements on deep generative models.

	\section*{Acknowledgments}
	This work was supported by BK21 Plus for Pioneers in Innovative Computing (Dept. of Computer Science and Engineering, SNU) funded by the National Research Foundation of Korea (NRF) (21A20151113068).
	We would also like to thank anonymous reviewers for their constructive feedbacks.	

	\bibliography{cslvm}
	\bibliographystyle{acl_natbib}
	
	\appendix
	\section{von Mises--Fisher Distribution}
	\label{app:vmf}
	A von Mises--Fisher (vMF) distribution is the distribution defined on a $m$-dimensional unit hypersphere.
	It is parameterized by two parameters: the mean direction $\bm\mu \in \mathbb{R}^m$ and the concentration $\kappa \in \mathbb{R}$.
	The probability density function (pdf) of $\text{vMF}(\bm\mu, \kappa)$ is defined by
	\begin{equation}
	\label{eqn:vmf-pdf}
	f(\mathbf{x};\bm\mu,\kappa) = C_m(\kappa) \exp (\kappa \bm\mu^\top \mathbf{x}),
	\end{equation}
	where
	\begin{equation*}
	C_m(\kappa) = \frac{\kappa^{m/2 - 1}}{(2\pi)^{m/2} I_{m/2 - 1} (\kappa)}
	\end{equation*}
	and $I_{v} (\kappa)$ is the modified Bessel function of the first kind at order $v$.
	Eq. \ref{eqn:vmf-pdf} is used in the computation of $\mathcal{R}^{\mathbf{z}}$.
	
	A sample from a vMF distribution is drawn from the acceptance-rejection scheme presented in Algorithm 1 of \citet{davidson2018svae}.
	In their algorithm, a stochastic variable obtained from the acceptance-rejection sampling does not depend on $\mathbf{\mu}$, thus the sampling process can be rewritten as a deterministic function that accepts the stochastic variable as input (i.e. reparameterization trick).
	
	The KL divergence between a vMF distribution $\text{vMF}(\bm\mu,\kappa)$ and the hyperspherical uniform distribution $\mathcal{U}(S^{m-1}) = \text{vMF}(\cdot,0)$ can be derived analytically:
	\begin{multline*}
	D_{KL}(\text{vMF}(\bm\mu, \kappa) \Vert \text{vMF}(\cdot, 0))
	\\ = \log C_m (\kappa) - \log \frac{\Gamma(m/2)}{2\pi^{m/2}} + \kappa \frac{I_{m/2}(\kappa)}{I_{m/2-1}(\kappa)}.
	\end{multline*}
	Note that the KL divergence does not depend on $\bm\mu$, thus the KL divergence is a constant if $\kappa$ is fixed.
	Intuitively, this is because the hyperspherical uniform distribution has equal probability density at every point on the unit hypersphere, and $D_{KL}(\text{vMF}(\bm\mu, \kappa) \Vert \text{vMF}(\cdot,0))$ should not be changed under rotations.
	
	\section{Derivation of Lower Bounds}
	\label{app:lb-derivation}
	Let $q_{\bm\theta} (\mathbf{z}_s | \cdot)$ be a distribution that has the same support with $p(\mathbf{z}_s)$.
	Then the KL divergence between $q_{\bm\theta} (\mathbf{z}_s | \cdot)$ and $p_{\bm\theta} (\mathbf{z}_s | \mathbf{x}_t, y)$ can be written as
	\begin{equation*}
	\begin{split}
	D&_{KL} (q_{\bm\phi}(\mathbf{z}_s|\cdot) \Vert p_{\bm\theta} (\mathbf{z}_s | \mathbf{x}_t, y))
	\\ &= \int q_{\bm\phi}(\mathbf{z}_s|\cdot) \log \frac{q_{\bm\phi}(\mathbf{z}_s|\cdot)}{p_{\bm\theta} (\mathbf{z}_s | \mathbf{x}_t, y)} d\mathbf{z}_s
	\\ &= \int q_{\bm\phi}(\mathbf{z}_s|\cdot) \log \frac{p_{\bm\theta}(\mathbf{x}_t,y)q_{\bm\phi}(\mathbf{z}_s|\cdot)}{p_{\bm\theta} (\mathbf{x}_t | \mathbf{z}_s, y) p(\mathbf{z}_s) p(y)} d\mathbf{z}_s
	\\ &= \log p_{\bm\theta} (\mathbf{x}_t, y) 
		+ D_{KL} (q_{\bm\phi} (\mathbf{z}_s | \cdot) \Vert p(\mathbf{z}_s))
		\\ & \qquad - \mathbb{E}_{q_{\bm\phi}(\mathbf{z}_s|\cdot)} [\log p_{\bm\theta} (\mathbf{x}_t | \mathbf{z}_s,y)] - \log p(y)
	\\ &\ge 0.
	\end{split}
	\end{equation*}
	From the above inequality we obtain the lower bound of $\log p_{\bm\theta} (\mathbf{x}_t,y)$ presented in Eq. \ref{eqn:elbo}.
	
	The lower bound of $\log p_{\bm\theta} (\mathbf{x}_t)$ (Eq. \ref{eqn:unsupervised-marginalization}) could be derived as follows.
	\begin{equation*}
	\begin{split}
	&\log p_{\bm\theta} (\mathbf{x}_t) = \log \sum_y \int p_{\bm\theta} (\mathbf{x}_t, \mathbf{z}_s, y) d\mathbf{z}_s \\
	&= \log \mathbb{E}_{q_{\bm\phi, \bm\psi} (y, \mathbf{z}_s|\mathbf{x}_s, \mathbf{x}_t)} \left[
		\frac{p_{\bm\theta} (\mathbf{x}_t|\mathbf{z}_s,y) p(\mathbf{z}_s) p(y)}
			{q_{\bm\phi, \bm\psi}(y, \mathbf{z}_s|\mathbf{x}_s, \mathbf{x}_t)}
	\right] \\
	&\ge \mathbb{E}_{q_{\bm\phi, \bm\psi} (y, \mathbf{z}_s|\mathbf{x}_s, \mathbf{x}_t)} \left[
		\log \frac{p_{\bm\theta} (\mathbf{x}_t|\mathbf{z}_s,y) p(\mathbf{z}_s) p(y)}
			{q_{\bm\phi, \bm\psi}(y, \mathbf{z}_s|\mathbf{x}_s, \mathbf{x}_t)}
	\right]
	\end{split}
	\end{equation*}
	From the graphical model $q_{\bm\phi, \bm\psi}(y, \mathbf{z}_s|\mathbf{x}_s, \mathbf{x}_t) = q_{\bm\phi} (\mathbf{z}_s|\mathbf{x}_s) q_{\bm\psi} (y|\mathbf{x}_s,\mathbf{x}_t)$, and thus
	\begin{equation*}
	\begin{split}
	&\mathbb{E}_{q_{\bm\phi, \bm\psi} (y, \mathbf{z}_s|\mathbf{x}_s, \mathbf{x}_t)} \left[
		\log \frac{p_{\bm\theta} (\mathbf{x}_t|\mathbf{z}_s,y) p(\mathbf{z}_s) p(y)}
			{q_{\bm\phi, \bm\psi}(y, \mathbf{z}_s|\mathbf{x}_s, \mathbf{x}_t)}
	\right] \\
	&= \!\begin{multlined}[t]\mathbb{E}_{q_{\bm\psi}} \left[
		\mathbb{E}_{q_{\bm\phi}} \left[
			\log \frac{p_{\bm\theta} (\mathbf{x}_t|\mathbf{z}_s,y) p(\mathbf{z}_s) p(y)}
			{q_{\bm\phi}(\mathbf{z}_s|\mathbf{x}_s)}
		\right]
	\right] \\
	- \mathbb{E}_{q_{\bm\psi}} [\log q_{\bm\psi} (y|\mathbf{x}_s,\mathbf{x}_t)]
	\end{multlined} \\
	&= \!\begin{multlined}[t]\mathbb{E}_{q_{\bm\psi}} \left[
		-\mathcal{L}_l^{gen}(\bm\theta, \bm\phi; \mathbf{x}_s,\mathbf{x}_t,y) \right] \\
		+ \mathcal{H} (q_{\bm\psi} (y|\mathbf{x}_s,\mathbf{x}_t))
	\end{multlined} \\
	&= -\mathcal{L}_u(\bm\theta, \bm\phi, \bm\psi; \mathbf{x}_s,\mathbf{x}_t).
	\end{split}
	\end{equation*}
	
	\section{Implementation Details}
	\label{app:implementation}
	We used PyTorch\footnote{\url{https://pytorch.org/}} and AllenNLP\footnote{\url{https://allennlp.org/}} libraries for implementation.
	The default weight initialization scheme of the AllenNLP library is used unless explicitly stated.
	
	For all CS-LVM experiments, the size of word embeddings and hidden dimensions of LSTMs are set to 300, and the size of label embeddings is 50.
	$g^{code}$ is implemented as a linear projection of the last hidden state of the encoder LSTM followed by normalization.
	$g^{out}$ is a linear projection followed by the softmax function, and we reuse the word embeddings as its weight matrix \cite{wolf2017tying,inan2017tying}.
	The discriminative classifier is a feedforward network with single hidden layer and the ReLU activation function, and the hidden dimension is set to 1200.
	We apply dropout on word embeddings and the classifier with probabilities $p_w$ and $p_c$ respectively.
	
	When multiple semantic constraints are used, to make uncertainty weights be always positive and be optimized stably, we instead use $\log \sigma_i^2$ as model parameter, as in \citet{kendall2018uncertainty}.
	Each $\log \sigma_i^2$ is initialized with zero.
	The temperature parameter of the Gumbel-Softmax is linearly annealed using the following schedule:
	\begin{equation*}
	\tau(t) = \max (0.1, 1.0 - rt),
	\end{equation*}
	where $r=10^{-4}$ is the annealing rate and $t$ is the training step.
	
	To find optimal hyperparameters, we performed a rough grid search on $\kappa \in \{100, 120, 150\}$, $\lambda \in \{0.2, 0.5, 0.8, 1.0\}$, $p_w \in \{0.25, 0.50, 0.75\}$, and $p_c \in \{0.1, 0.2\}$.
	The KL divergence between a posterior and the prior is 23.57, 27.09, 31.60 when $\kappa$ is set to 100, 120, 150 respectively.
	
	For the LSTM-vMF-VAE experiments, we used the same hyperparameters and grid search scheme with those of the CS-LVM, except that we perform an additional search on the dimension of a latent code $d\in \{50, 150, 300\}$.
	
	Adam optimizer \cite{kingma2015adam} with learning rate $\gamma=10^{-3}$ is used for all experiments, except for 1k QQP experiments where stochastic gradient descent optimizer is used.
	When fine-tuning the model, we set $\gamma$ to $10^{-4}$.
	For other hyperparameters, we follow the configuration suggested by the authors.	
	Best hyperparameter configurations found for SNLI and QQP datasets are presented in Tables \ref{table:snli-hyperparams} and \ref{table:qqp-hyperparams}.
	
	\begin{table}
		\small
		\centering
		\begin{tabular}{l r r r r}
			\hline
			\bf Model & $\kappa$ & $\lambda$ & $p_w$ & $p_c$ \\
			\hline
			28k & 150 & 0.8 & 0.75 & 0.1 \\
			59k & 100 & 1.0 & 0.75 & 0.1 \\
			120k & 120 & 0.8 & 0.50 & 0.1 \\
			\hline
		\end{tabular}
		\caption{Hyperparameters for the SNLI models.}
		\label{table:snli-hyperparams}
	\end{table}
	
	\begin{table}
		\small
		\centering
		\begin{tabular}{l r r r r}
			\hline
			\bf Model & $\kappa$ & $\lambda$ & $p_w$ & $p_c$ \\
			\hline
			1k & 100 & 0.8 & 0.50 & 0.2 \\
			5k & 120 & 0.5 & 0.75 & 0.2 \\
			10k & 150 & 0.5 & 0.75 & 0.1 \\
			25k & 100 & 0.5 & 0.75 & 0.1 \\
			\hline
		\end{tabular}
		\caption{Hyperparameters for the QQP models.}
		\label{table:qqp-hyperparams}
	\end{table}

	\section{Generated Examples}
	\label{app:generated-examples}
	We used beam search with $B=10$ when generating sentences, and the length normalization \cite{wu2016gnmt} is applied with $\alpha=0.7$.
	
	Examples are presented in Tables \ref{table:samples-base}--\ref{table:samples-mu}.
	Though almost all generated hypotheses are realistic, we see that they lack diversity and fail to encode label information in some cases.
	For example, the phrase `is/are sleeping' appears in generated sentences frequently when conditioned on the `contradiction' label, likely because generating a set of simple patterns could be a shortcut to the objective.
	In Table \ref{table:eval-gen-datasets}, we verified from experiments that adding constraints helps enhancing accuracy and diversity, however a model is still relatively in favor of generating `easy' sentences.
	We conjecture that the problem has its root in the fact that the primary objective of our model is to correctly classify the input, not to generate diverse outputs.
	
	\begin{table*}[htb]
		\centering
		\scriptsize
		\begin{tabular}{p{0.26\textwidth}|p{0.2\textwidth}|p{0.2\textwidth}|p{0.2\textwidth}}
			\hline
			\textbf{Input} & \textbf{Entailment} & \textbf{Neutral} & \textbf{Contradiction} \\
			\hline
			little kids enjoy sprinklers by running through them outdoors .
			& kids are running .
			& the children are siblings .
			& the children are playing video games . \\
			\hline
			blurry people walking in the city at night .
			& people are walking outside .
			& the people are going to work .
			& the people are inside . \\
			\hline
			a woman sits in a chair under a tree and plays an acoustic guitar .
			& a woman is playing an instrument .
			& the woman is a musician .
			& a woman is playing the flute . \\
			\hline
			three men converse in a crowd .
			& three men are talking .
			& \sout{the men are talking .}
			& the men are sleeping . \\
			\hline
			a woman in a yellow shirt seated at a table .
			& a woman is sitting .
			& \sout{a woman is sitting at a table .}
			& the woman is standing . \\
			\hline
			a woman hugs a fluffy white dog .
			& a woman is holding a dog .
			& \sout{a woman is with her dog .}
			& a woman is sleeping . \\
			\hline
			a crowd of people in colorful dresses .
			& people in costumes
			& the people are in a parade .
			& \sout{the people are sitting in a circle .} \\
			\hline
			a clown making a balloon animal for a pretty lady .
			& \sout{a clown is entertaining a crowd .}
			& the clown is entertaining a crowd .
			& the clown is sleeping . \\
			\hline \\
		\end{tabular}
		\caption{
			Sentences generated from the CS-LVM model trained on the SNLI dataset.
			Failure cases are denoted by \sout{strikethrough} text.
		}
		\label{table:samples-base}
	\end{table*}
	
	\begin{table*}[htb]
		\centering
		\scriptsize
		\begin{tabular}{p{0.26\textwidth}|p{0.2\textwidth}|p{0.2\textwidth}|p{0.2\textwidth}}
			\hline
			\textbf{Input} & \textbf{Entailment} & \textbf{Neutral} & \textbf{Contradiction} \\
			\hline
			little kids enjoy sprinklers by running through them outdoors .
			& kids are playing outside .
			& the kids are playing in the water .
			& the kids are sleeping . \\
			\hline
			blurry people walking in the city at night .
			& people are walking .
			& the people are walking to work .
			& the people are inside . \\
			\hline
			a woman sits in a chair under a tree and plays an acoustic guitar .
			& a woman is playing music .
			& the woman is a musician .
			& a woman is sleeping . \\
			\hline
			three men converse in a crowd .
			& three men are talking .
			& three men are talking about politics .
			& the men are sleeping . \\
			\hline
			a woman in a yellow shirt seated at a table .
			& a woman is sitting .
			& a tall human sitting .
			& the woman is standing . \\
			\hline
			a woman hugs a fluffy white dog .
			& a woman is holding a dog .
			& the dog belongs to the woman .
			& the dog is black . \\
			\hline
			a crowd of people in colorful dresses .
			& people in costumes
			& the people are in a parade .
			& the people are sleeping . \\
			\hline
			a clown making a balloon animal for a pretty lady .
			& a clown is performing .
			& the clown is entertaining a crowd .
			& the clown is sleeping . \\
			\hline
		\end{tabular}
		\caption{
			Sentences generated from the CS-LVM + $\mathcal{R}^y$ model trained on the SNLI dataset.
			Note that failed examples in Table \ref{table:samples-base} are corrected due to the use of $\mathcal{R}^y$.
		}
		\label{table:samples-y}
	\end{table*}
	
	\begin{table*}[htb]
		\centering
		\scriptsize
		\begin{tabular}{p{0.26\textwidth}|p{0.2\textwidth}|p{0.2\textwidth}|p{0.2\textwidth}}
			\hline
			\textbf{Input} & \textbf{Entailment} & \textbf{Neutral} & \textbf{Contradiction} \\
			\hline
			little kids enjoy sprinklers by running through them outdoors .
			& \sout{kids are playing in water .}
			& the kids are having fun .
			& the kids are sleeping . \\
			\hline
			blurry people walking in the city at night .
			& people are walking .
			& the people are walking to work .
			& the people are inside . \\
			\hline
			a woman sits in a chair under a tree and plays an acoustic guitar .
			& a woman is playing an instrument .
			& the woman is a musician .
			& a woman is playing the drums . \\
			\hline
			three men converse in a crowd .
			& three men are talking .
			& three men are talking about politics .
			& the men are sleeping . \\
			\hline
			a woman in a yellow shirt seated at a table .
			& a woman is sitting .
			& \sout{a woman is sitting at a table .}
			& the woman is standing \\
			\hline
			a woman hugs a fluffy white dog .
			& a woman is holding a dog .
			& a woman is playing with her dog .
			& a woman is sleeping . \\
			\hline
			a crowd of people in colorful dresses .
			& people are wearing costumes .
			& the people are in a parade .
			& \sout{the people are sitting down .} \\
			\hline
			a clown making a balloon animal for a pretty lady .
			& a clown performs .
			& \sout{the clown is a clown .}
			& the clown is sleeping . \\
			\hline
		\end{tabular}
		\caption{
			Sentences generated from the CS-LVM + $\mathcal{R}^{\mathbf{z}}$ model trained on the SNLI dataset.
			Failure cases are denoted by \sout{strikethrough} text.
		}
		\label{table:samples-z}
	\end{table*}
	
	\begin{table*}[htb]
		\centering
		\scriptsize
		\begin{tabular}{p{0.26\textwidth}|p{0.2\textwidth}|p{0.2\textwidth}|p{0.2\textwidth}}
			\hline
			\textbf{Input} & \textbf{Entailment} & \textbf{Neutral} & \textbf{Contradiction} \\
			\hline
			little kids enjoy sprinklers by running through them outdoors .
			& kids are playing outside .
			& the kids are having fun .
			& the kids are sleeping . \\
			\hline
			blurry people walking in the city at night .
			& people are walking .
			& the people are walking to work .
			& the people are inside . \\
			\hline
			a woman sits in a chair under a tree and plays an acoustic guitar .
			& a woman is playing an instrument .
			& the woman is a musician .
			& a woman is playing the piano . \\
			\hline
			three men converse in a crowd .
			& three men are talking .
			& three men are talking about politics .
			& the men are sleeping . \\
			\hline
			a woman in a yellow shirt seated at a table .
			& a woman is sitting .
			& \sout{a woman is sitting at a table .}
			& the woman is standing \\
			\hline
			a woman hugs a fluffy white dog .
			& a woman is holding a dog .
			& the dog belongs to the woman .
			& a woman is petting a cat . \\
			\hline
			a crowd of people in colorful dresses .
			& people are dressed up .
			& the people are in a parade .
			& \sout{the people are sitting down .} \\
			\hline
			a clown making a balloon animal for a pretty lady .
			& \sout{a clown is blowing bubbles .}
			& \sout{the clown is a clown .}
			& the clown is sleeping . \\
			\hline
		\end{tabular}
		\caption{
			Sentences generated from the CS-LVM + $\mathcal{R}^{\bm\mu}$ model trained on the SNLI dataset.
			Failure cases are denoted by \sout{strikethrough} text.
		}
		\label{table:samples-mu}
	\end{table*}

\end{document}